%% file: main.tex
\documentclass[10pt,twocolumn,letterpaper]{article}

\usepackage[accsupp]{axessibility}

\usepackage{cvpr}              

\input{preamble}

\definecolor{cvprblue}{rgb}{0.21,0.49,0.74}
\usepackage[pagebackref,breaklinks,colorlinks,citecolor=cvprblue]{hyperref}

\def\paperID{4190} 
\def\confName{CVPR}
\def\confYear{2024}

\title{Condition-Aware Neural Network for Controlled Image Generation}

\author{
    Han Cai$^{1*}$, Muyang Li$^1$, Zhuoyang Zhang$^2$, Qinsheng Zhang$^3$, Ming-Yu Liu$^3$, Song Han$^{1,3}$ \\
    $^1$MIT, $^2$Tsinghua University, $^3$NVIDIA  \\
    \url{https://github.com/mit-han-lab/efficientvit}
}

\begin{document}
\maketitle
\let\thefootnote\relax\footnotetext{$^*$Work done during an internship at NVIDIA.}

\input{sec/0_abstract}
\input{sec/1_intro}
\input{sec/3_method}
\input{sec/4_exp}
\input{sec/2_related}
\input{sec/5_end}
{
    \small
    \bibliographystyle{unsrt}
    \bibliography{main}
}

\end{document}

%% file: preamble.tex
\usepackage[dvipsnames]{xcolor}

\usepackage{graphicx}
\usepackage{amsmath,amsfonts,amsthm,amssymb}
\usepackage{booktabs}
\usepackage{fancyhdr}
\usepackage{float,wrapfig}
\usepackage{pifont}

\newcommand{\xmark}{\ding{55}}%

\usepackage[utf8]{inputenc}
\usepackage[T1]{fontenc}

\usepackage{epsfig}

\usepackage{adjustbox}
\usepackage{array}
\usepackage{colortbl}
\usepackage{hhline}
\usepackage{multirow}

\usepackage{bm}
\usepackage{nicefrac}
\usepackage{microtype}

\usepackage{changepage}
\usepackage{extramarks}
\usepackage{lastpage}
\usepackage{setspace}
\usepackage{soul}
\usepackage{xspace}
\usepackage{indentfirst}
\usepackage{cuted}

\usepackage{algorithm,algorithmic}
\usepackage{enumitem}

\usepackage{wasysym}
\usepackage{duckuments}

%% file: sec/0_abstract.tex
\begin{abstract}
We present \textbf{\underline{C}ondition-\underline{A}ware Neural \underline{N}etwork (CAN)}, a new method for adding control to image generative models. In parallel to prior conditional control methods, CAN controls the image generation process by dynamically manipulating the weight of the neural network. This is achieved by introducing a condition-aware weight generation module that generates conditional weight for convolution/linear layers based on the input condition. We test CAN on class-conditional image generation on ImageNet and text-to-image generation on COCO. CAN consistently delivers significant improvements for diffusion transformer models, including DiT and UViT. In particular, \underline{CA}N combined with EfficientVi\underline{T} (CaT) achieves 2.78 FID on ImageNet 512$\times$512, surpassing DiT-XL/2 while requiring 52$\times$ fewer MACs per sampling step. 
\end{abstract}

%% file: sec/1_intro.tex
\section{Introduction}
\label{sec:intro}
Large-scale image \cite{rombach2022high,kang2023scaling,balaji2022ediffi,saharia2022photorealistic} and video generative models \cite{videoworldsimulators2024,blattmann2023stable} have demonstrated astounding capacity in synthesizing photorealistic images and videos. To convert these models into productive tools for humans, a critical step is adding control. Instead of letting the model randomly generate data samples, we want the generative model to follow our instructions (e.g., class label, text, pose) \cite{zhang2023adding}. 

Extensive studies have been conducted to achieve this goal. For example, in GANs \cite{brock2018large,karras2019style}, a widespread solution is to use adaptive normalization \cite{huang2017arbitrary,perez2018film} that dynamically scales and shifts the intermediate feature maps according to the input condition. In addition, another widely used technique is to use cross-attention \cite{rombach2022high} or self-attention \cite{bao2022all} to fuse the condition feature with the image feature. Though differing in the used operations, these methods share the same underlying mechanism, i.e., adding control by feature space manipulation. Meanwhile, the neural network weight (convolution/linear layers) remains the same for different conditions. 

\begin{figure}[t]
    \centering
    \includegraphics[width=\linewidth]{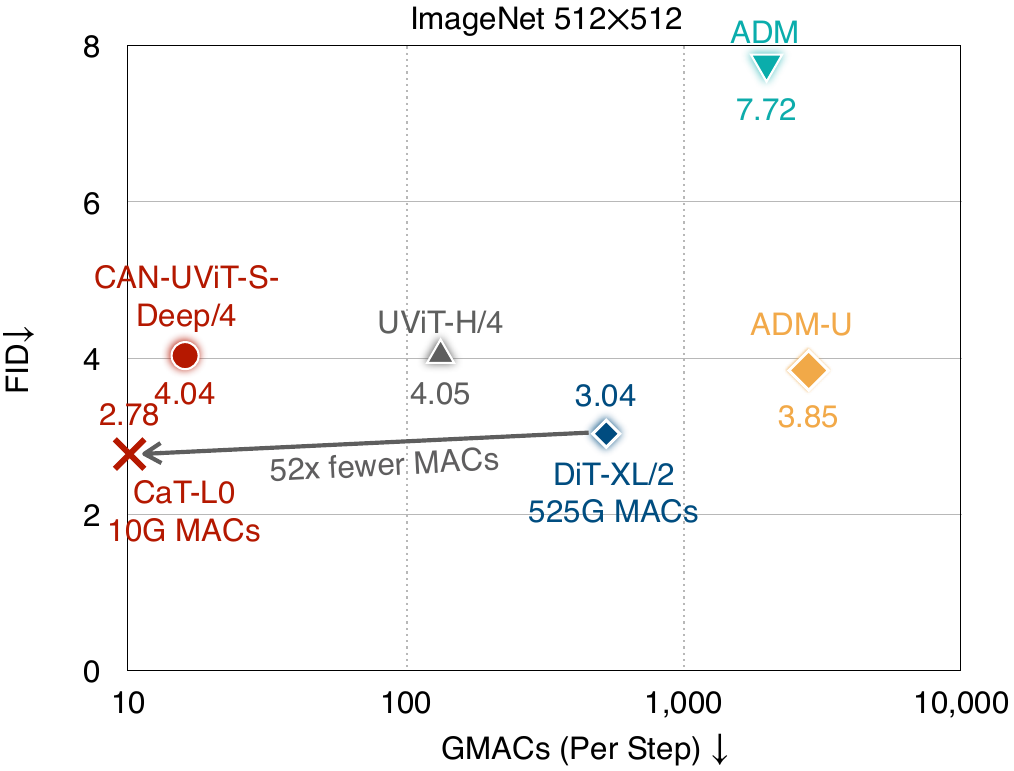}
    \caption{\textbf{Comparing CAN Models and Prior Image Generative Models on ImageNet 512$\times$512.} With the new conditional control method, we significantly improve the performance of controlled image generative models. Combining CAN and EfficientViT \cite{cai2023efficientvit}, our CaT model provides 52$\times$ MACs reduction per sampling step than DiT-XL/2 \cite{peebles2023scalable} without performance loss.}
    \label{fig:teaser}
\end{figure}

\begin{figure*}[t]
    \centering
    \includegraphics[width=\linewidth]{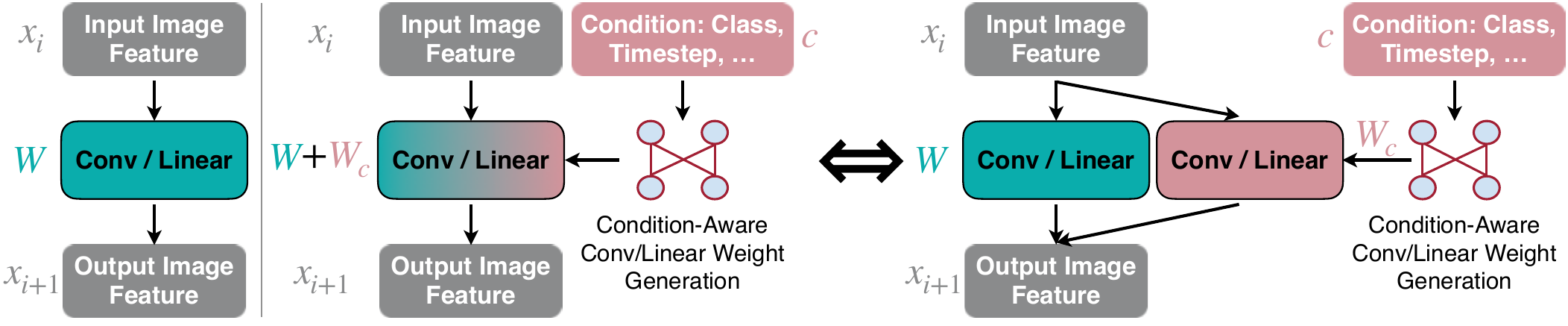}
    \caption{\textbf{Illustration of Condition-Aware Neural Network.} \emph{Left:} A regular neural network with static convolution/linear layers. \emph{Right:} A condition-aware neural network and its equivalent form. }
    \label{fig:method}
\end{figure*}

This work aims to answer the following questions: \emph{i) Can we control image generative models by manipulating their weight? ii) Can controlled image generative models benefit from this new conditional control method?} 

To this end, we introduce \textbf{\underline{C}ondition-\underline{A}ware Neural \underline{N}etwork (CAN)}, a new conditional control method based on weight space manipulation. Differentiating from a regular neural network, CAN introduces an additional weight generation module (Figure~\ref{fig:method}). The input to this module is the condition embedding, which consists of the user instruction (e.g., class label) and the timestep for diffusion models \cite{ho2020denoising}. The module's output is the conditional weight used to adapt the static weight of the convolution/linear layer. We conduct extensive ablation study experiments investigating the practical use of CAN on diffusion transformers. Our study reveals two critical insights for CAN. First, rather than making all layers condition-aware, we find carefully choosing a subset of modules to be condition-aware (Figure~\ref{fig:method_arch}) is beneficial for both efficiency and performance (Table~\ref{tab:ablation_model_varients}). Second, we find directly generating the conditional weight is much more effective than adaptively merging a set of base static layers \cite{yang2019condconv} for conditional control (Figure~\ref{fig:ablation}). 

We evaluate CAN on two representative diffusion transformer models, including DiT \cite{peebles2023scalable}, and UViT \cite{bao2022all}. CAN achieves significant performance boosts for all these diffusion transformer models while incurring negligible computational cost increase (Figure~\ref{fig:can_diffusion}). We also find that CAN alone provides effective conditional control for image generative models, delivering lower FID and higher CLIP scores than prior conditional control methods (Table~\ref{tab:vs_ada_norm_and_att}). Apart from applying CAN to existing diffusion transformer models, we further build a new family of diffusion transformer models called \textbf{CaT} by marrying CAN and EfficientViT \cite{cai2023efficientvit} (Figure~\ref{fig:cat}).
We summarize our contributions as follows:
\begin{itemize}[leftmargin=*]
\item We introduce a new mechanism for controlling image generative models. To the best of our knowledge, our work is the first to demonstrate the effectiveness of weight manipulation for conditional control.
\item We propose Condition-Aware Neural Network, a new conditional control method for controlled image generation. We also provide design insights to make CAN usable in practice. 
\item Our CAN consistently improves performances on image generative models, outperforming prior conditional control methods by a significant margin. In addition, CAN can also benefit the deployment of image generative models. Achieving a better FID on ImageNet 512$\times$512, our CAN model requires 52$\times$ fewer MACs than DiT-XL/2 per sampling step (Figure~\ref{fig:teaser}), paving the way for diffusion model applications on edge devices. 
\end{itemize}

%% file: sec/3_method.tex
\section{Method}
\label{sec:method}

\begin{figure*}[t]
    \centering
    \includegraphics[width=0.9\linewidth]{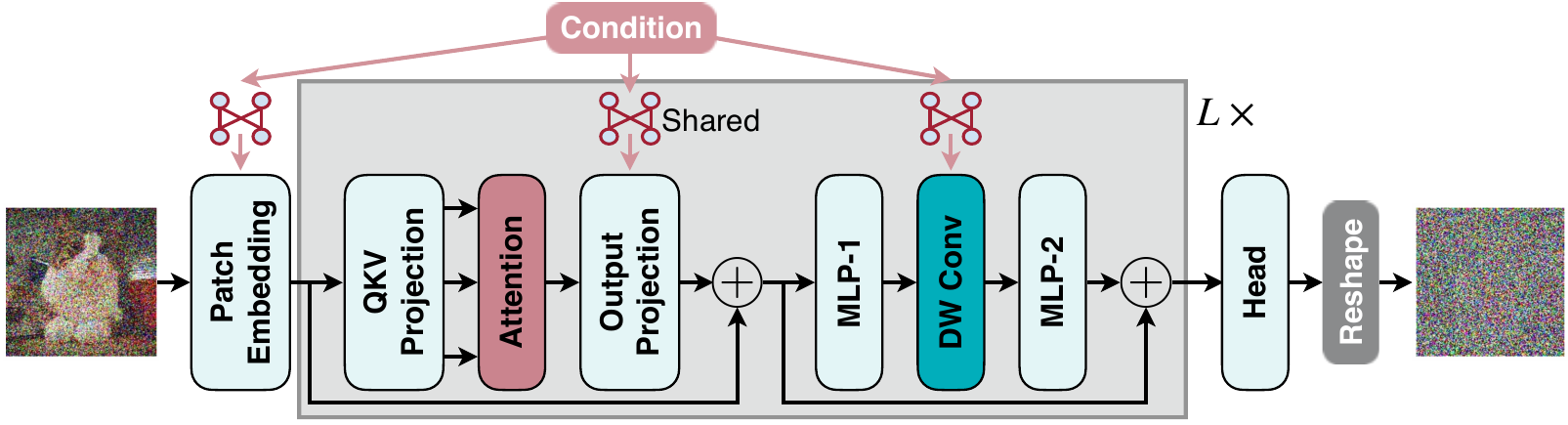}
    \caption{\textbf{Overview of Applying CAN to Diffusion Transformer.} The patch embedding layer, the output projection layers in self-attention, and the depthwise convolution (DW Conv) layers are condition-aware. The other layers are static. All output projection layers share the same conditional weight while still having their own static weights.}
    \label{fig:method_arch}
\end{figure*}

\subsection{Condition-Aware Neural Network}

The image generation process can be viewed as a mapping from the source domain (noise or noisy image) to the target domain (real image). For controlled image generation, the target data distribution is different given different conditions (e.g., cat images' data distribution vs. castle images' data distribution). In addition, the input data distribution is also different for diffusion models \cite{ho2020denoising} at different timesteps. Despite these differences, prior models use the same static convolution/linear layers for all cases, limiting the overall performance due to negative transfer between different sub-tasks \cite{wu2020understanding}. To alleviate this issue, one possible solution is to have an expert model \cite{shazeer2017outrageously} for each sub-task. However, this approach is infeasible for practical use because of the enormous cost. Our condition-aware neural network (CAN) tackles this issue by enabling the neural network to adjust its weight dynamically according to the given condition, instead of explicitly having the expert models. 

Figure~\ref{fig:method} demonstrates the general idea of CAN. The key difference from a regular neural network is that CAN has an extra conditional weight generation module. This module takes the condition embedding $c$ as the input and outputs the conditional weight $W_c$. In addition to the conditional weight $W_c$, each layer has the static weight $W$. During training and inference, $W_c$ and $W$ are fused into a single kernel call by summing the weight values. This is equivalent to applying $W_c$ and $W$ independently on the input image feature and then adding their outputs. 

\subsection{Practical Design}\label{sec:practical_design}
\paragraph{Which Modules to be Condition-Aware?} 
Theoretically, we can make all layers in the neural network condition-aware. However, in practice, this might not be a good design. First, from the performance perspective, having too many condition-aware layers might make the model optimization challenging. Second, from the efficiency perspective, while the computational overhead of generating the conditional weight for all layers is negligible${^1}$\footnote{$^1$It is because the sequence length (or spatial size) of the condition embedding is much smaller than the image feature.}, it will incur a significant parameter overhead. For example, let's denote the dimension of the condition embedding as $d$ (e.g., 384, 512, 1024, etc) and the model's static parameter size as $\#$params. Using a single linear layer to map from the condition embedding to the conditional weight requires $\#$params $\times d$ parameters, which is impractical for real-world use. In this work, we carefully choose a subset of modules to apply CAN to solve this issue. 

An overview of applying CAN to diffusion transformer \cite{peebles2023scalable,bao2022all} is provided in Figure~\ref{fig:method_arch}. Depthwise convolution \cite{chollet2017xception} has a much smaller parameter size than regular convolution, making it a low-cost candidate to be condition-aware. Therefore, we add a depthwise convolution in the middle of FFN following the previous design \cite{cai2023efficientvit}. We conduct ablation study experiments on ImageNet 256$\times$256 using UViT-S/2 \cite{bao2022all} to determine the set of modules to be condition-aware. \emph{All the models, including the baseline model, have the same architecture. The only distinction is the set of condition-aware modules is different.} 

We summarize the results in Table~\ref{tab:ablation_model_varients}. We have the following observations in our ablation study experiments:
\begin{itemize}[leftmargin=*]
\item Making a module condition-aware does not always improve the performance. For example, using a static head gives a lower FID and a higher CLIP score than using a condition-aware head (row \#2 vs. row \#4 in Table~\ref{tab:ablation_model_varients}). 
\item Making depthwise convolution layers, the patch embedding layer, and the output projection layers condition-aware brings a significant performance boost. It improves the FID from 28.32 to 8.82 and the CLIP score from 30.09 to 31.74.  
\end{itemize}
Based on these results, we chose this design for CAN. Details are illustrated in Figure~\ref{fig:method_arch}. For the depthwise convolution layers and the patch embedding layer, we use a separate conditional weight generation module for each layer, as their parameter size is small. In contrast, we use a shared conditional weight generation module for the output projection layers, as their parameter size is large. Since different output projection layers have different static weights, we still have different weights for different output projection layers. 

\begin{table}[t]
\small\centering
\begin{tabular}{l | c c }
\toprule
\multicolumn{3}{l}{\textbf{ImageNet 256$\times$256, UViT-S/2}} \\
\midrule
Models & FID $\downarrow$ & CLIP Score $\uparrow$ \\
\midrule
1. Baseline (Static Conv/Linear) & 28.32 & 30.09 \\
\midrule
\multicolumn{3}{l}{\emph{Making Modules Condition-Aware:}} \\
\midrule
2. DW Conv & 11.18 & 31.54 \\
\midrule
3. + Patch Embedding & 10.23 & 31.61 \\
\textcolor{gray}{4. or + Head (\xmark)} & \textcolor{gray}{12.29} & \textcolor{gray}{31.40} \\
\midrule
5. + Output Projection & 8.82 & 31.74 \\
\textcolor{gray}{6. or + QKV Projection (\xmark)} & \textcolor{gray}{9.71} & \textcolor{gray}{31.66} \\
\textcolor{gray}{7. or + MLP (\xmark)} & \textcolor{gray}{10.06} & \textcolor{gray}{31.62} \\ 
\bottomrule
\end{tabular}
\caption{\textbf{Ablation Study on Making Which Modules Condition-Aware.}}
\label{tab:ablation_model_varients}
\end{table}

\begin{figure}[t]
    \centering
    \includegraphics[width=0.7\linewidth]{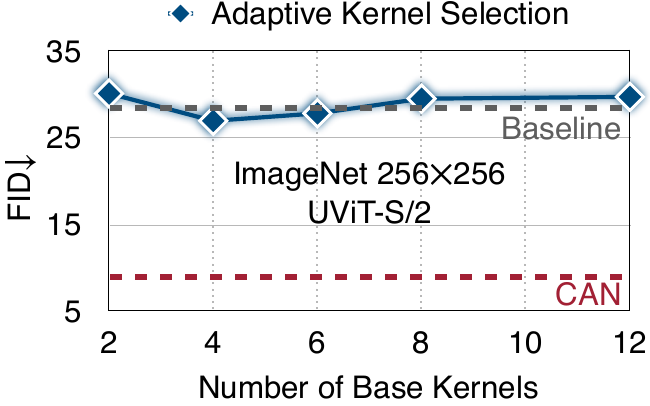}
    \caption{\textbf{CAN is More Effective than Adaptive Kernel Selection.}}
    \label{fig:ablation}
\end{figure}

\begin{figure*}[t]
    \centering
    \includegraphics[width=\linewidth]{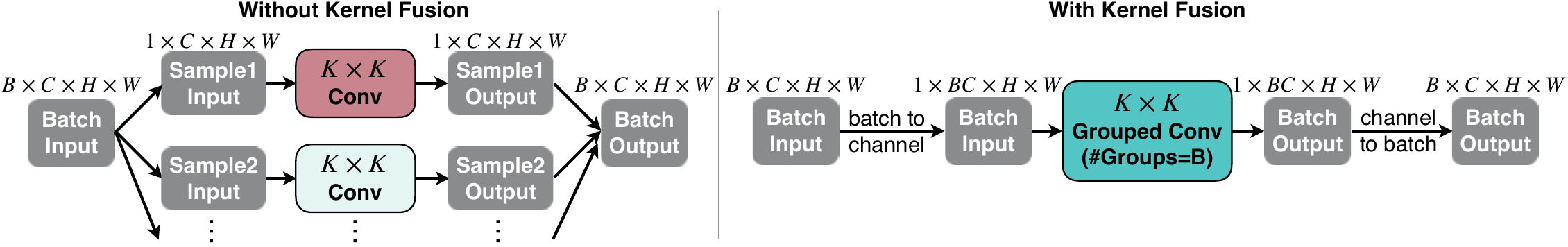}
    \caption{\textbf{Practical Implementation of CAN.} \emph{Left:} The condition-aware layers have different weights for different samples. A naive implementation requires running the kernel call independently for each sample, which incurs a large overhead for training and batch inference. \emph{Right:} An efficient implementation for CAN. We fuse all kernel calls into a grouped convolution. We insert a batch-to-channel transformation before the kernel call and add a channel-to-batch conversion after the kernel call to preserve the functionality.}
    \label{fig:method_train}
\end{figure*}

\begin{figure*}[t]
    \centering
    \includegraphics[width=\linewidth]{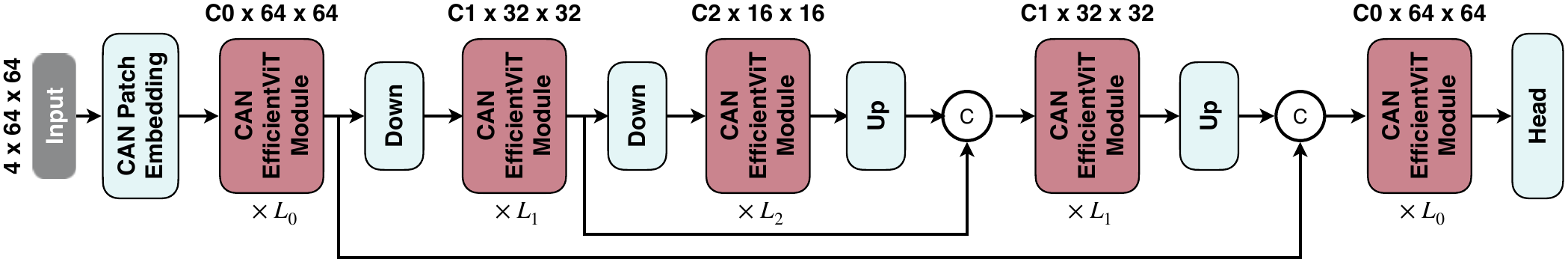}
    \caption{\textbf{Macro Architecture of CaT.} Benefiting from EfficientViT's linear computational complexity \cite{cai2023efficientvit}, we can keep the high-resolution stages without efficiency concerns.}
    \label{fig:cat}
\end{figure*}

\begin{figure*}[t]
    \centering
    \includegraphics[width=\linewidth]{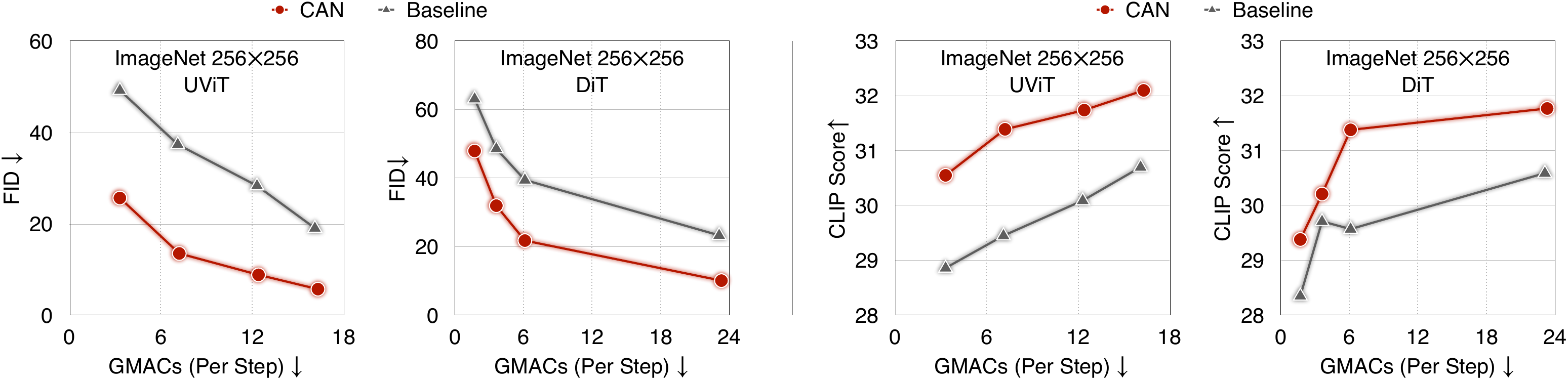}
    \caption{\textbf{CAN Results on Different UViT and DiT Variants.} CAN consistently delivers lower FID and higher CLIP score for UViT and DiT variants. }
    \label{fig:can_diffusion}
\end{figure*}

\paragraph{CAN vs. Adaptive Kernel Selection.} Instead of directly generating the conditional weight, another possible approach is maintaining a set of base convolution kernels and dynamically generating scaling parameters to combine these base kernels \cite{kang2023scaling,yang2019condconv}. This approach's parameter overhead is smaller than CAN. However, this adaptive kernel selection strategy cannot match CAN's performance (Figure~\ref{fig:ablation}). It suggests that dynamic parameterization alone is not the key to better performances; better condition-aware adaptation capacity is critical. 

\paragraph{Implementation.} Since the condition-aware layers have different weights given different samples, we cannot do the batch training and inference. Instead, we must run the kernel calls independently for each sample, as shown in Figure~\ref{fig:method_train} (left). This will significantly slow down the training process on GPU. To address this issue, we employ an efficient implementation for CAN (Figure~\ref{fig:method_train} right). The core insight is to fuse all convolution kernel calls \cite{karras2020analyzing} into a grouped convolution where $\#$Groups is the batch size $B$. We do the batch-to-channel conversion before running the grouped convolution to preserve the functionality. After the operation, we add the channel-to-batch transformation to convert the feature map to the original format.

Theoretically, with this efficient implementation, there will be negligible training overhead compared to running a static model. In practice, as NVIDIA GPU supports regular convolution much better than grouped convolution, we still observe 30\%-40\% training overhead. This issue can be addressed by writing customized CUDA kernels. We leave it to future work.

%% file: sec/4_exp.tex
\section{Experiments}
\label{sec:exp}

\subsection{Setups}
\paragraph{Datasets.} Due to resource constraints, we conduct class-conditional image generation experiments using the ImageNet dataset and use COCO for text-to-image generation experiments. For large-scale text-to-image experiments \cite{shi2020improving}, we leave them to future work. 

\paragraph{Evaluation Metric.} Following the common practice, we use FID \cite{heusel2017gans} as the evaluation metric for image quality. In addition, we use the CLIP score \cite{hessel2021clipscore} as the metric for controllability. We use the public CLIP ViT-B/32 \cite{radford2021learning} for measuring the CLIP score, following \cite{shi2020improving}. The text prompts are constructed following CLIP's zero-shot image classification setting \cite{radford2021learning}. 

\paragraph{Implementation Details.} We apply CAN to recent diffusion transformer models, including DiT \cite{peebles2023scalable} and UViT \cite{bao2022all}. We follow the training setting suggested in the official paper or GitHub repository. By default, classifier-free guidance \cite{ho2022classifier} is used for all models unless explicitly stated. The baseline models' architectures are the same as the CAN models', having depthwise convolution in FFN layers. We implement our models using Pytorch and train them using A6000 GPUs. Automatic mixed-precision is used during training. In addition to applying CAN to existing models, we also build a new family of diffusion transformers called \textbf{CaT} by marrying CAN and EfficientViT \cite{cai2023efficientvit}. The macro architecture of CaT is illustrated in Figure~\ref{fig:cat}.

\subsection{Ablation Study}
We train all models for 80 epochs with batch size 1024 (around 100K iterations) for ablation study experiments unless stated explicitly. All models use DPM-Solver \cite{lu2022dpm} with 50 steps for sampling images.

\paragraph{Effectiveness of CAN.} Figure~\ref{fig:can_diffusion} summarizes the results of CAN on various UViT and DiT variants. CAN significantly improves the image quality and controllability over the baseline for all variants. Additionally, these improvements come with negligible computational cost overhead. Therefore, CAN also enhances efficiency by delivering the same FID and CLIP score with lower-cost models. 

\begin{figure}[t]
    \centering
    \includegraphics[width=\linewidth]{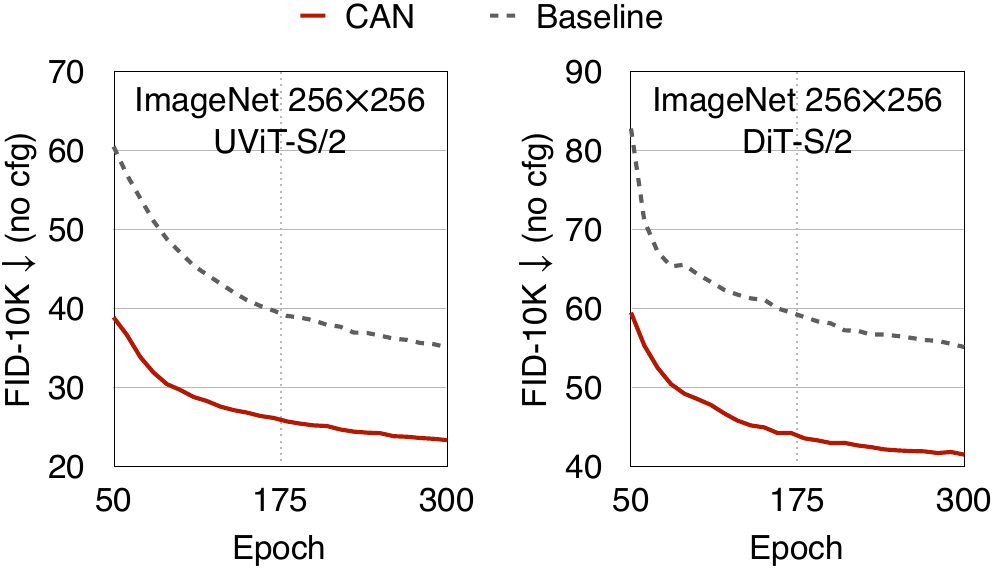}
    \caption{\textbf{Training Curve.} CAN's improvements are not due to faster convergence. We observe consistent FID improvements when trained longer. }
    \label{fig:training_curve}
\end{figure}

Figure~\ref{fig:training_curve} compares the training curves of CAN and baseline on UViT-S/2 and DiT-S/2. We can see that the absolute improvement remains significant when trained longer for both models. It shows that the improvements are not due to faster convergence. Instead, adding CAN improves the performance upper bound of the models. 

\paragraph{Analysis.} For diffusion models, the condition embedding contains both the class label and timestep. To dissect which one is more important for the conditional weight generation process, we conduct the ablation study experiments using UViT-S/2, and summarize the results in Table~\ref{tab:analysis}. We find that:
\begin{itemize}[leftmargin=*]
\item The class label information is more important than the timestep information in the weight generation process. Adding class label alone provides 5.15 lower FID and 0.33 higher CLIP score than adding timestep alone. 
\item Including the class label and the timestep in the condition embedding delivers the best results. Therefore, we stick to this design in the following experiments. 
\end{itemize}

\begin{table}[t]
\small\centering
\begin{tabular}{l | c c }
\toprule
\multicolumn{3}{l}{\textbf{ImageNet 256$\times$256, UViT-S/2}} \\
\midrule
Models & FID $\downarrow$ & CLIP Score $\uparrow$ \\
\midrule
Baseline & 28.32 & 30.09 \\
\midrule
CAN (Timestep Only) & 15.16 & 31.26 \\
CAN (Class Label Only) & 10.01 & 31.59 \\
\midrule
CAN (All) & 8.82 & 31.74 \\
\bottomrule
\end{tabular}
\caption{\textbf{Ablation Study on the Effect of Each Condition for CAN.} }
\label{tab:analysis}
\end{table}

\begin{table}[t]
\small\centering\setlength{\tabcolsep}{3pt}
\begin{tabular}{l | c c }
\toprule
\multicolumn{3}{l}{\textbf{ImageNet 256$\times$256, DiT-S/2}} \\
\midrule
Models & FID $\downarrow$ & CLIP Score $\uparrow$ \\
\midrule
Adaptive Normalization & 39.44 & 29.57 \\
\midrule
CAN Only & 26.44 & 30.54\\
CAN + Adaptive Normalization & 21.76 & 30.86 \\ 
\bottomrule
\toprule
\multicolumn{3}{l}{\textbf{ImageNet 256$\times$256, UViT-S/2}} \\
\midrule
Models & FID $\downarrow$ & CLIP Score $\uparrow$ \\
\midrule
Attention (Condition as Tokens) & 28.32 & 30.09 \\
\midrule
CAN Only & 8.79 & 31.75 \\
CAN + Attention (Condition as Tokens) & 8.82 & 31.74 \\ 
\bottomrule
\end{tabular}
\caption{\textbf{Comparison with Prior Conditional Control Methods.} CAN can work alone without adding other conditional control methods.}
\label{tab:vs_ada_norm_and_att}
\end{table}

\begin{table*}[t]
\small\centering
\begin{tabular}{l | c c | c c | c }
\toprule
\multicolumn{6}{l}{\textbf{ImageNet 512$\times$512}} \\
\midrule
Models & FID-50K (no cfg) $\downarrow$ & FID-50K $\downarrow$ & \#MACs (Per Step) $\downarrow$ & \#Steps $\downarrow$ & \#Params $\downarrow$ \\
\midrule
ADM \cite{dhariwal2021diffusion} & 23.24 & 7.72 & 1983G & 250 & 559M \\
ADM-U \cite{dhariwal2021diffusion} & \textbf{9.96} & 3.85 & 2813G & 250 & 730M \\
\midrule
UViT-L/4 \cite{bao2022all} & - & 4.67 & 77G & 50 & 287M \\
UViT-H/4 \cite{bao2022all} & - & 4.05 & 133G & 50 & 501M \\
DiT-XL/2 \cite{peebles2023scalable} & 12.03 & 3.04 & 525G & 250 & 675M \\
\midrule
\textbf{CAN (UViT-S-Deep/4)} & 23.40 & 4.04 & 16G & 50 & 185M \\
\midrule
\textbf{CaT-L0} & 14.25 & 2.78 & 10G & 20 & 377M \\
\textbf{CaT-L1} & 10.64 & \textbf{2.48} & 12G & 20 & 486M \\
\bottomrule
\toprule
\multicolumn{6}{l}{\textbf{ImageNet 256$\times$256}} \\
\midrule
Models & FID-50K (no cfg) $\downarrow$ & FID-50K $\downarrow$ & \#MACs (Per Step) $\downarrow$ & \#Steps $\downarrow$ & \#Params $\downarrow$ \\
\midrule
LDM-4 \cite{rombach2022high} & 10.56 & 3.60 & - & 250 & 400M \\
\midrule
UViT-L/2 \cite{bao2022all} & - & 3.40 & 77G & 50 & 287M \\
UViT-H/2 \cite{bao2022all} & - & 2.29 & 133G & 50 & 501M \\
DiT-XL/2 \cite{peebles2023scalable} & 9.62 & 2.27 & 119G & 250 & 675M \\
\midrule
\textbf{CAN (UViT-S/2)} & 16.20 & 3.52 & 12G & 50 & 147M \\
\textbf{CAN (UViT-S-Deep/2)} & 11.89 & 2.78 & 16G & 50 & 182M \\
\midrule
\textbf{CaT-B0} & \textbf{8.81} & \textbf{2.09} & 12G & 30 & 475M \\
\bottomrule
\end{tabular}
\caption{\textbf{Class-Conditional Image Generation Results on ImageNet.}}
\label{tab:imagenet_sota}
\end{table*}

\paragraph{Comparison with Prior Conditional Control Methods.} In prior experiments, we kept the original conditional control methods of DiT (adaptive normalization) and UViT (attention with condition as tokens) unchanged while adding CAN. To see if CAN can work alone and the comparison between CAN and previous conditional control methods, we conduct experiments and provide the results in Table~\ref{tab:vs_ada_norm_and_att}. We have the following findings:
\begin{itemize}[leftmargin=*]
\item CAN alone can work as an effective conditional control method. For example, CAN alone achieves a 13.00 better FID and 0.97 higher CLIP score than adaptive normalization on DiT-S/2. In addition, CAN alone achieves a 19.53 lower FID and 1.66 higher CLIP score than attention (condition as tokens) on UViT-S/2. 
\item CAN can be combined with other conditional control methods to achieve better results. For instance, combining CAN with adaptive normalization provides the best results for DiT-S/2. 
\item For UViT models, combining CAN with attention (condition as tokens) slightly hurts the performance. Therefore, we switch to using CAN alone on UViT models in the following experiments. 
\end{itemize}

\subsection{Comparison with State-of-the-Art Models}
We compare our final models with other diffusion models on ImageNet and COCO. The results are summarized in Table~\ref{tab:imagenet_sota} and Table~\ref{tab:coco_sota}. For CaT models, we use UniPC \cite{zhao2024unipc} for sampling images to reduce the number of sampling steps. 

\paragraph{Class-Conditional Generation on ImageNet 256$\times$256.} As shown in Table~\ref{tab:imagenet_sota} (bottom), with the classifier-free guidance (cfg), our CaT-B0 achieves 2.09 FID on ImageNet 256$\times$256, outperforming DiT-XL/2 and UViT-H/2. More importantly, our CaT-B0 is much more compute-efficient than these models: 9.9$\times$ fewer MACs than DiT-XL/2 and 11.1$\times$ fewer MACs than UViT-H/2. Without the classifier-free guidance, our CaT-B0 also achieves the lowest FID among all compared models (8.81 vs. 9.62 vs. 10.56).  

\paragraph{Class-Conditional Generation on ImageNet 512$\times$512.} On the more challenging 512$\times$512 image generation task, we observe the merits of CAN become more significant. For example, our CAN (UViT-S-Deep/4) can match the performance of UViT-H (4.04 vs. 4.05) while only requiring 12\% of UViT-H's computational cost per diffusion step. Additionally, our CaT-L0 delivers 2.78 FID on ImageNet 512$\times$512, outperforming DiT-XL/2 (3.04 FID) that requires 52$\times$ higher computational cost per diffusion step. In addition, by slightly scaling up the model, our CaT-L1 further improves the FID from 2.78 to 2.48.

\begin{table}[t]
\small\centering\setlength{\tabcolsep}{3pt}
\begin{tabular}{l | c c | c }
\toprule
\multicolumn{4}{l}{\textbf{ImageNet 512$\times$512}} \\
\midrule
Models & \#Steps $\downarrow$ & Orin Latency $\downarrow$ & FID $\downarrow$ \\
\midrule
DiT-XL/2 \cite{peebles2023scalable} & 250 & 45.9s & 3.04 \\
\midrule
\textbf{CaT-L0} & 20 & 0.2s & 2.78 \\
\bottomrule
\end{tabular}
\caption{\textbf{NVIDIA Jetson AGX Orin Latency vs. FID.} Latency is profiled with TensorRT and fp16.}
\label{tab:orin_latency}
\end{table}

In addition to computational cost comparisons, Table~\ref{tab:orin_latency} compares CaT-L0 and DiT-XL/2 on NVIDIA Jetson AGX Orin. The latency is measured with TensorRT, fp16. Delivering a better FID on ImageNet 512$\times$512, CaT-L0 combined with a training-free fast sampling method (UniPC) runs 229$\times$ faster than DiT-XL/2 on Orin. It is possible to further push the efficiency frontier by combining CaT with training-based few-step methods \cite{salimans2022progressive,song2023consistency}, showing the potential of enabling real-time diffusion model applications on edge devices. 

\begin{figure*}[t]
    \centering
    \includegraphics[width=\linewidth]{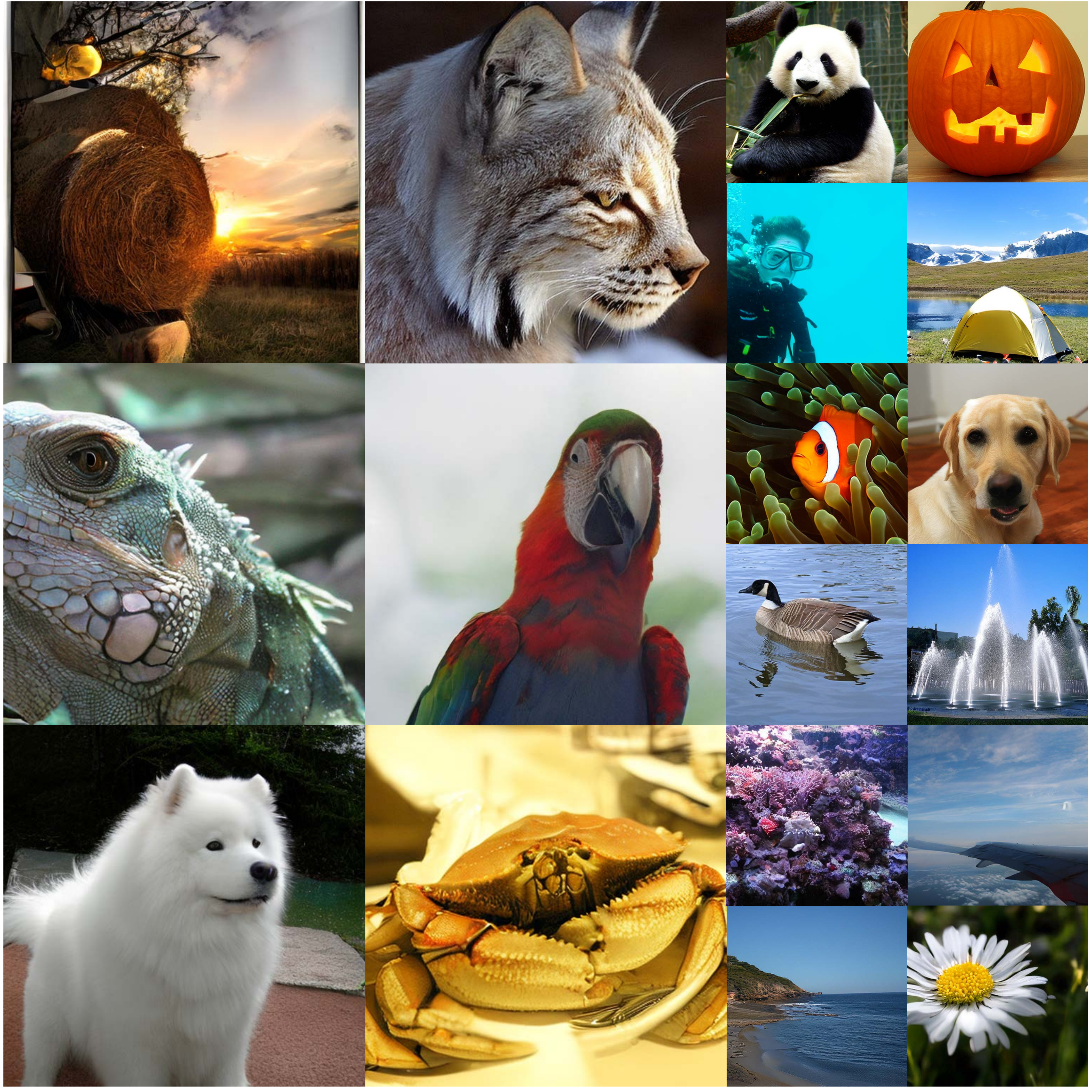}
    \caption{\textbf{Samples of Generated Images by CAN Models.}}
    \label{fig:samples}
\end{figure*}

Apart from quantitative results, Figure~\ref{fig:samples} provides samples from the randomly generated images by CAN models, which demonstrate the capability of our models in generating high-quality images. 

\paragraph{Text-to-Image Generation on COCO 256$\times$256.} For text-to-image generation experiments on COCO, we follow the same setting used in UViT \cite{bao2022all}. Specifically, models are trained from scratch on the COCO 2014 training set. Following UViT \cite{bao2022all}, we randomly sample 30K text prompts from the COCO 2014 validation set to generate images and then compute FID. We use the same CLIP encoder as in UViT for encoding the text prompts. The results are summarized in Table~\ref{tab:coco_sota}. Our CaT-S0 achieves a similar FID as UViT-S-Deep/2 while having much less computational cost (19G MACs $\rightarrow$ 3G MACs). It justifies the generalization ability of our models. 

\begin{table}[t]
\small\centering
\begin{tabular}{l | c | c c }
\toprule
\multicolumn{4}{l}{\textbf{COCO 256$\times$256}} \\
\midrule
Models & FID-30K $\downarrow$ & \#MACs $\downarrow$ & \#Params $\downarrow$ \\
\midrule
Friro & 8.97 & - & 512M \\
\midrule
UViT-S/2 & 5.95 & 15G & 45M \\
UViT-S-Deep/2 & 5.48 & 19G & 58M \\
\bottomrule
\textbf{CaT-S0} & 5.49 & 3G & 169M \\
\textbf{CaT-S1} & \textbf{5.22} & 7G & 307M \\
\bottomrule
\end{tabular}
\caption{\textbf{Text-to-Image Generation Results on COCO 256$\times$256.} }
\label{tab:coco_sota}
\end{table}

%% file: sec/2_related.tex
\section{Related Work}
\label{sec:related}

\paragraph{Controlled Image Generation.} Controlled image generation requires the models to incorporate the condition information into the computation process to generate related images. Various techniques have been developed in the community for controlled image generation. One typical example is adaptive normalization \cite{perez2018film} that regresses scale and shift parameters from the condition information and applies the feature-wise affine transformation to influence the output. Apart from adaptive normalization, another typical approach is to treat condition information as tokens and use either cross-attention \cite{rombach2022high} or self-attention \cite{bao2022all} to fuse the condition information. ControlNet \cite{zhang2023adding} is another representative technique that uses feature-wise addition to add extra control to pre-trained text-to-image diffusion models. In parallel to these techniques, this work explores another mechanism for adding conditional control to image generative models, i.e., making the weight of neural network layers (conv/linear) condition-aware.  

\paragraph{Dynamic Neural Network.} Our work can be viewed as a new type of dynamic neural network. Apart from adding conditional control explored in this work, dynamically adapting the neural network can be applied to many deep learning applications. For example, CondConv \cite{yang2019condconv} proposes to dynamically combine a set of base convolution kernels based on the input image feature to increase the model capacity. Similarly, the mixture-of-expert \cite{shazeer2017outrageously} technique uses a gating network to route the input to different experts dynamically. For efficient deployment, once-for-all network \cite{cai2019once} and slimmable neural network \cite{yu2018slimmable} dynamically adjust the neural network architecture according to the given efficiency constraint to achieve better tradeoff between efficiency and accuracy. 

\paragraph{Weight Generating Networks.} Our conditional weight generation module can be viewed as a new kind of weight generation network specially designed for adding conditional control to generative models. There are some prior works exploiting weight generation networks in other scenarios. For example, \cite{ha2017hypernetworks} proposes to use a small network to generate weights
for a larger network. These weights are the same for every example in the dataset for better parameter efficiency. Additionally, weight generation networks have been applied to neural architecture search to predict the weight of a neural network given its architecture \cite{brock2017smash} to reduce the training and search cost \cite{cai2018proxylessnas} of neural architecture search. 

\paragraph{Efficient Deep Learning Computing.} Our work is also connected to efficient deep learning computing \cite{han2015deep,cai2021network} that aims to improve the efficiency of deep learning models to make them friendly for deployment on hardware. State-of-the-art image generative models \cite{rombach2022high,kang2023scaling,shi2020improving,saharia2022photorealistic} have enormous computation and memory costs, which makes it challenging to deploy them on resource-constrained edge devices while maintaining high quality. Our work can improve the efficiency of the controlled generative models by delivering the same performance with fewer diffusion steps and lower-cost models. For future work, we will explore combining our work and efficient deep learning computing techniques \cite{li2023q,cai2019once} to futher boost efficiency. 

%% file: sec/5_end.tex
\section{Conclusion}
\label{sec:conclusion}
In this work, we studied adding control to image generative models by manipulating their weight. We introduced a new conditional control method, called \textbf{\underline{C}ondition-\underline{A}ware Neural \underline{N}etwork (CAN)}, and provided efficient and practical designs for CAN to make it usable in practice. We conducted extensive experiments on class-conditional generation using ImageNet and text-to-image generation using COCO to evaluate CAN's effectiveness. CAN delivered consistent and significant improvements over prior conditional control methods. We also built a new family of diffusion transformer models by marrying CAN and EfficientViT. For future work, we will apply CAN to more challenging tasks like large-scale text-to-image generation, video generation, etc. 

\subsection*{Acknowledgments}
\noindent This work is supported by MIT-IBM Watson AI Lab, Amazon, MIT Science Hub, and National Science Foundation.